\documentclass[runningheads]{llncs}

\usepackage{eccv}

\usepackage{eccvabbrv}

\usepackage{graphicx}
\usepackage{booktabs}
\usepackage{threeparttable}
\usepackage{multicol}
\usepackage{multirow}
\usepackage{colortbl}
\usepackage{tcolorbox}
\usepackage{xcolor}
\definecolor{darkred}{rgb}{0.8,0,0}

\usepackage[accsupp]{axessibility}  
\usepackage[breaklinks,colorlinks,citecolor=eccvblue]{hyperref}

\usepackage{orcidlink}

\begin{document}

\title{SkeletonAgent: An Agentic Interaction Framework for Skeleton-based Action Recognition}

\titlerunning{SkeletonAgent}

\author{
Hongda Liu\inst{1,2}
\orcidlink{0000-0002-4644-6222} 
\and
Yunfan Liu\inst{2}$^{\star}$
\orcidlink{0000-0001-8929-4866} 
\and
Changlu Wang\inst{1,2}
\orcidlink{0009-0005-5995-2510} 
\and
Yunlong Wang\inst{1}
\orcidlink{0000-0002-3535-308X} 
\and
Zhenan Sun\inst{1}\thanks{Corresponding authors.}
\orcidlink{0000-0003-4029-9935}
}

\authorrunning{H. Liu et al.}

\institute{
NLPR, Institute of Automation, Chinese Academy of Sciences
\and
University of Chinese Academy of Sciences
\\
\email{\{hongda.liu,changlu.wang,yunlong.wang\}@cripac.ia.ac.cn, liuyunfan@ucas.ac.cn, znsun@nlpr.ia.ac.cn}
}

\maketitle

\begin{abstract}
Recent advances in skeleton-based action recognition increasingly leverage semantic priors from Large Language Models (LLMs) to enrich skeletal representations.
However, the LLM is typically queried in isolation from the recognition model and receives no performance feedback.
As a result, it often fails to deliver the targeted discriminative cues critical to distinguish similar actions.
To overcome these limitations, we propose SkeletonAgent, a novel framework that bridges the recognition model and the LLM through two cooperative agents, \ie, \textit{Questioner} and \textit{Selector}.
Specifically, the \textit{Questioner} identifies the most frequently confused classes and supplies them to the LLM as context for more targeted guidance.
Conversely, the \textit{Selector} parses the LLM’s response to extract precise joint-level constraints and feeds them back to the recognizer, enabling finer-grained cross-modal alignment.
Comprehensive evaluations on five benchmarks, including NTU RGB+D, NTU RGB+D 120, Kinetics-Skeleton, FineGYM, and UAV-Human, demonstrate that SkeletonAgent consistently outperforms state-of-the-art benchmark methods.
The code is available at \url{https://github.com/firework8/SkeletonAgent}.
\end{abstract}

\section{Introduction}

Skeleton-based action recognition has received sustained attention for its robustness in complex environments and computational efficiency~\cite{song2017end,shi2019two,do2024skateformer}.
Considering the skeletal structure, most approaches employ Graph Convolutional Networks (GCNs) to model inter-joint dependencies and capture spatiotemporal dynamics~\cite{yan2018spatial,zhou2024blockgcn}.
Building on recent advances in Large Language Models (LLMs), a growing line of research seeks to enrich skeletal representations with multi-modal semantic priors, thereby enhancing the recognition accuracy~\cite{xiang2023generative,sato2023prompt,xu2025language,kuang2025zero}.

\begin{figure}[t]
\centering
\includegraphics[width=\linewidth]{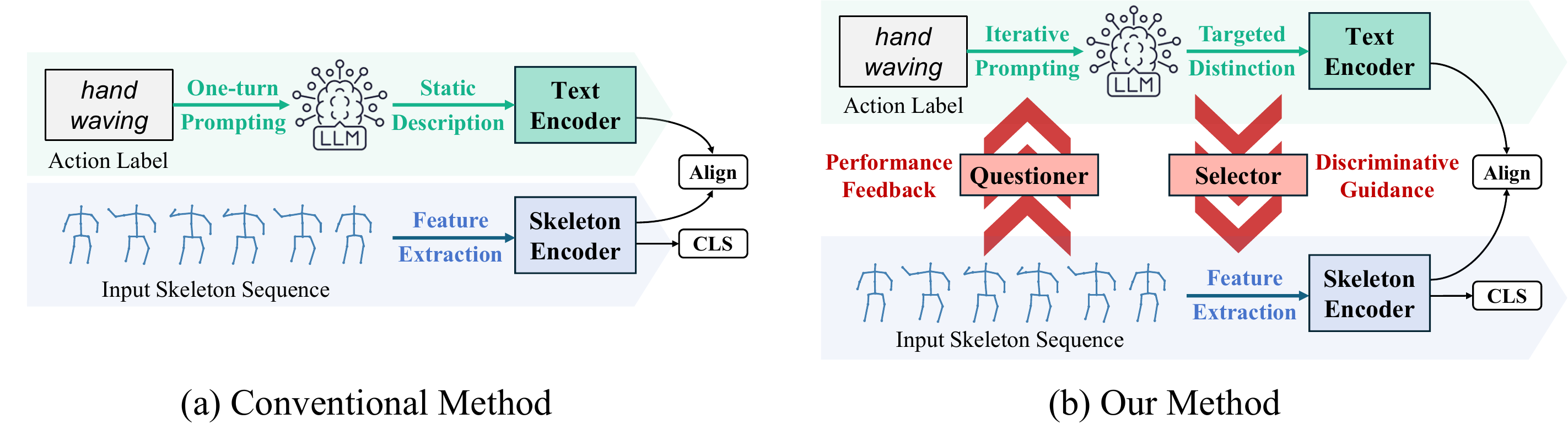}
\setlength{\abovecaptionskip}{-8pt}
\setlength{\belowcaptionskip}{0pt}
\caption{ 
Comparison of (a) conventional multi-modal skeleton-based action recognition pipeline and (b) our SkeletonAgent.
SkeletonAgent incorporates two cooperative agents, \ie, \textit{Questioner} and \textit{Selector}, that create an interactive bridge to facilitate information flow between the action recognizer and the guidance provider (LLM), thereby narrowing the modality gap.
}
\label{fig:diagram_diff}
\end{figure}

To be specific, these methods typically construct an associated space to bridge the modality gap between skeletal and textual semantics~\cite{sato2023prompt,li2024sa,zhu2024part}. 
As illustrated in \cref{fig:diagram_diff} (a), they feed the LLM with prompts containing action labels, and then transform the generated descriptions into text features. 
The subsequent alignment establishes robust cross-modal relationships between these embeddings.
Consequently, the language-derived vector enriches the skeletal representation with the LLM’s semantic priors.

Despite the remarkable progress, these multi-modal approaches still struggle to distinguish classes with similar joint trajectories (\ie, \textit{similar actions}).
The core limitation lies in \textbf{the absence of an interactive feedback loop} between the action recognizer and the guidance provider (LLM).
Concretely, this disconnect undermines discrimination in two ways:
\textbf{(1) Homogeneous textual semantics.}
For similar actions, the text descriptions produced by existing single-turn prompts are often semantically alike.
This homogeneity conceals key discriminative cues, resulting in semantics that lack the class-specific nuances needed for accurate recognition.
\textbf{(2) Coarse Cross-Modal Alignment.}
Even if the text contains differences, the holistic aligning paradigm adopted by most methods lacks fine-grained control over the correspondence between textual fragments and specific motion patterns.
This often misplaces skeletal features relative to their intended semantics, preventing recognition models from attending to critical subtle distinctions.

To address these limitations, we propose SkeletonAgent, a novel framework for skeleton-based action recognition.
The core of SkeletonAgent lies in the interactive feedback mechanism between recognition models and language models.
Inspired by recent advances in agents, we leverage this agent-based framework with online execution capabilities, providing the LLM with real-time performance feedback and steering the recognizer toward discriminative patterns.

Specifically, the interaction framework involves two bridging agent components, namely the \textit{Questioner} and the \textit{Selector}.
As illustrated in \cref{fig:diagram_diff} (b), the \textit{Questioner} supplies the LLM with instantaneous context-aware feedback from recognition models (\eg, the current challenging actions), prompting the LLM to emphasize subtle class distinctions in its responses. 
Then, the \textit{Selector} iteratively queries and summarizes critical joints and distinguishing characteristics in the conversation history, converting them into local constraints and global semantics that can be leveraged by recognition models.
Unlike previous studies that seek static details in a single turn, our two agents engage in an iterative and adaptive collaboration with recognition models, enabling them to focus on more distinctive cues.
Experimental results demonstrate that SkeletonAgent establishes new state-of-the-art performance, achieving 94.5\% accuracy on the NTU RGB+D dataset and thereby validating the effectiveness of the proposed approach.

The main contributions are summarized as follows:
\begin{itemize}
  \item We introduce SkeletonAgent, a novel agent-based framework that establishes the online interaction between the recognition model and LLM, enabling targeted guidance for fine-grained action discrimination. 

  \item We design a bidirectional feedback mechanism, where two specialized agents collaborate to identify and emphasize key distinctive cues, thereby achieving precise cross-modal alignment.
  
  \item Extensive experiments demonstrate that SkeletonAgent consistently achieves state-of-the-art performance on five benchmarks, including NTU RGB+D, NTU RGB+D 120, Kinetics-Skeleton, FineGYM, and UAV-Human.
\end{itemize}

\section{Related Work}

\subsection{Skeleton-based Action Recognition}

Skeleton-based action recognition is to classify actions from sequences of estimated human key points.
Early pioneering methods employ recurrent neural networks (RNNs) and convolutional neural networks (CNNs) to model the skeletons~\cite{du2015hierarchical,li2017skeleton,duan2022revisiting}.
Following ST-GCN~\cite{yan2018spatial}, the mainstream works in recent years have gradually focused on models based on Graph Convolutional Networks (GCNs)~\cite{shi2019two,duan2022dg,zhang2024modular,liu2025revealing,chang2025hierarchical}.
To enhance the power of GCNs, various follow-up studies have explored more flexible modeling of spatial relations and temporal correlations~\cite{chen2021channel,chi2022infogcn,lee2023hierarchically,zhou2025adaptive}.
In addition, with the popularity of vision transformer, transformer-based methods have been investigated for skeleton data~\cite{xin2023skeleton,wu2024frequency,do2024skateformer}.
Regarding the challenges of similar actions, recent studies have also explored diverse solutions, such as shifting viewpoints~\cite{hou2022shifting}, amplifying key details~\cite{liu2025revealing}, and enhancing decoupled features~\cite{huang2023graph,zhou2023learning}.

Despite these advances, existing methods remain largely uni-modal, lacking targeted semantic guidance to distinguish fine-grained motions. 
In contrast, we introduce an online multi-modal training paradigm that leverages LLMs to provide context-aware discriminative cues. 
Through the interaction process, our method achieves fine-grained alignment to improve the skeletal representations.

\subsection{LLM Agents}

An agent is defined as an entity that makes decisions and executes tasks in a dynamic, real-time environment to achieve specific goals.
Recent advances in LLMs, particularly their emerging reasoning and planning capabilities~\cite{zhou2022least,shinn2023reflexion}, have inspired recent research in natural language processing to leverage agents for complex tasks.
The integration of chain-of-thought reasoning and self-correction could further extend the functionalities of LLMs~\cite{wei2022chain}.
These agent-based approaches have achieved notable success across various scenarios, such as prompt engineering, online search, and tool utilization~\cite{chang2024agentboard,yao2022webshop}. 
Simultaneously, the vision community has also adopted LLM agents for tasks like long video understanding and text-to-image generation~\cite{wang2024videoagent,yang2025vca,liao2025motionagent}.

Concretely, a key strength of LLM agents is their interactive capability, \ie, the ability to process and utilize information across the real-time environment to make informed decisions. 
Inspired by this perspective, we propose SkeletonAgent, a novel framework that departs from prior methods by explicitly modeling the interactive relationship between recognition models and language models.
Our agents enable dynamic interactions that progressively refine discriminative skeletal representations.
Through this structured collaboration and selective inspiration, our method could achieve fine-grained discrimination.

\subsection{Multi-modal Representation Learning}

Recent advances in language models have greatly accelerated progress in multi-modal representation learning.
Methods such as CLIP~\cite{radford2021learning} and ALIGN~\cite{jia2021scaling} demonstrate that vision–language training can learn robust representations for downstream tasks.
For instance, ActionCLIP~\cite{wang2023actionclip} follows this training scheme and extends alignment to the video domain for action recognition. 
Beyond vision and language, ULIP~\cite{xue2023ulip} learns a unified representation space incorporating point clouds to improve understanding.
Similarly, MotionCLIP~\cite{tevet2022motionclip} aligns human motion manifolds with CLIP latent space for human action generation. 
In the skeleton-based action recognition task, GAP~\cite{xiang2023generative} introduces generative prompts and a multi-part contrastive loss to align skeleton sequences with textual descriptions.
Additionally, PURLS~\cite{zhu2024part} explores visual-semantic alignment at different local and global scales.
Following that, Neuron~\cite{chen2025neuron} leverages the side information with diverse granularity to refine skeletal features.

However, these methods primarily rely on static text descriptions, which are often homogeneous across similar actions and lack fine-grained control over alignment. 
In contrast, our framework addresses this via the dynamic interaction mechanism.
The proposed \textit{Questioner} generates targeted distinctions based on recognition feedback, while the \textit{Selector} injects salient joints and semantics into the model, enabling precise and adaptive cross-modal alignment.

\section{Method}

\begin{figure}[t]
\centering
\includegraphics[width=\linewidth]{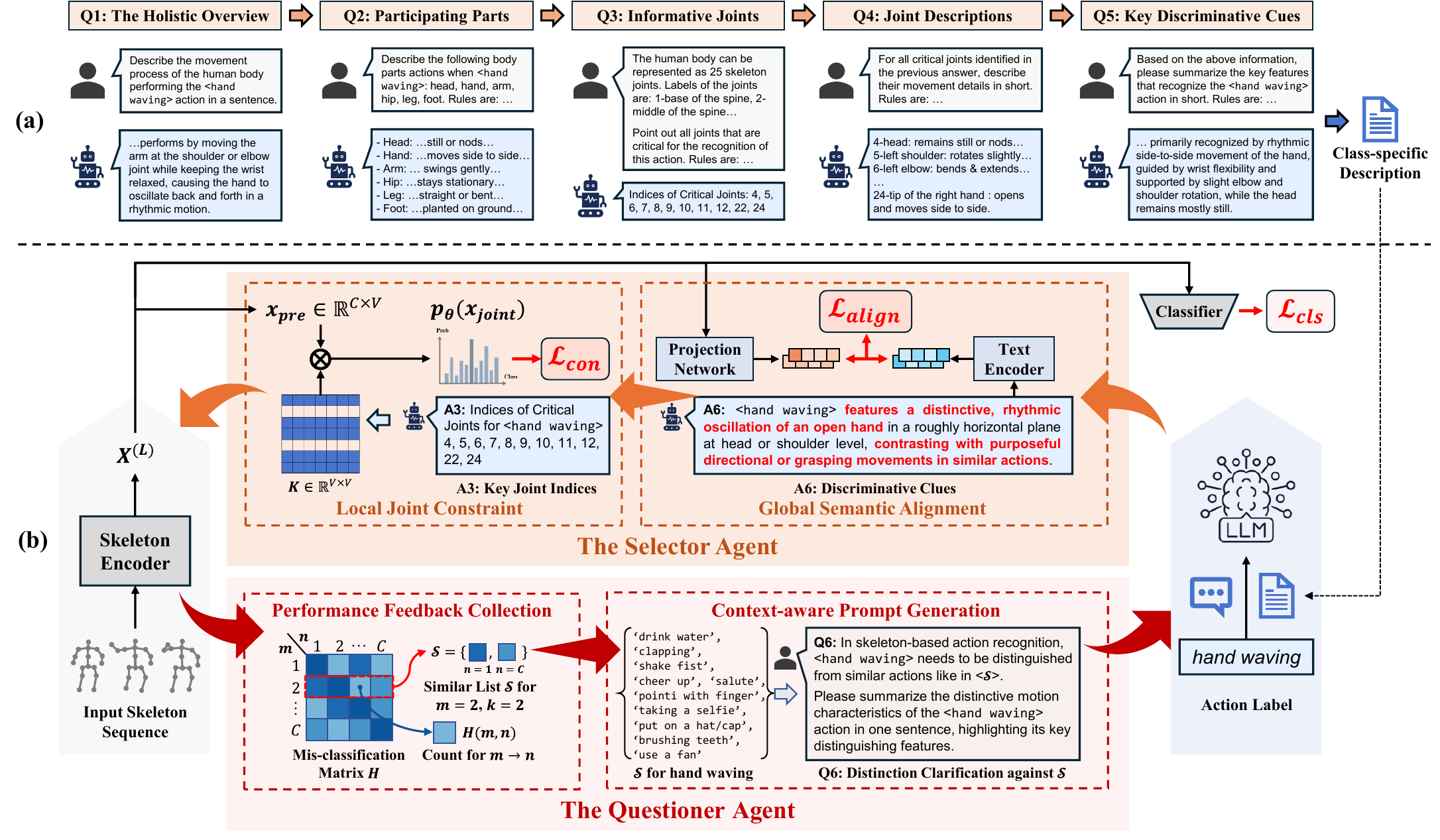}
\setlength{\abovecaptionskip}{0pt}
\setlength{\belowcaptionskip}{0pt}
\caption{ 
Illustration of (a) the generation pipeline of class-specific description, and (b) the overall framework of SkeletonAgent.
}
\label{fig:figure2}
\end{figure}

In this section, we introduce SkeletonAgent, a novel interaction framework for skeleton-based action recognition.
We first briefly formalize the task and establish notation, followed by an overview of the architecture.
Then, we detail the motivation and implementation of the two interactive agents, \ie, the \textit{Questioner} and the \textit{Selector}. 
A schematic overview of SkeletonAgent is provided in \cref{fig:figure2}.

\subsection{Preliminaries}
\label{sec:3_1}

For skeleton-based action recognition, an input sequence with $T$ frames is expressed as a tensor $\mathbf{X} \in \mathbb{R}^{C \times T \times V}$, where $V$ denotes the number of joints and $C$ denotes the latent dimension.
Due to its inherent structural characteristics, the graph-like human skeleton is usually processed with GCNs.
Let $\mathbf{X}^{(l)}$ be the feature at the $l$-th layer, the graph convolution can be formulated as
\begin{equation}
  \mathbf{X}^{(l)} = \sigma (\mathbf{D}^{-\frac{1}{2}}\mathbf{A}^{(l)}\mathbf{D}^{\frac{1}{2}}\mathbf{X}^{(l-1)}\mathbf{W}^{(l)})
  \label{eq:graph_conv}
\end{equation}
where $\mathbf{A}$ is the adjacency matrix representing joint connectivity, $\mathbf{D}\in\mathbb{R}^{N\times N}$ is the normalized degree matrix, $\mathbf{W}^{(l)}$ denotes the learnable parameter of the $l$-th layer, and $\sigma$ indicates the ReLU function.
With a network comprising $L$ layers with parameters $\theta$, the final feature $\mathbf{X}^{(L)}$ is passed to a classifier that produces one-hot encoding, which is trained using the cross-entropy loss $\mathcal{L}_{cls}$ as follows
\begin{equation}
  \mathcal{L}_{cls} = - \mathbf{y} \log \mathbf{p_{\theta}(x)} 
  \label{eq:ce_loss}
\end{equation}
where $\mathbf{y}$ is the one-hot ground-truth action label, and $\mathbf{p_{\theta}(x)}$ denotes the predicted probability distribution.

\subsection{Architecture Overview}
\label{sec:3_2}

The proposed SkeletonAgent establishes the dynamic interaction between the action recognizer and language models. 
As shown in Fig.~\ref{fig:figure2}, the framework consists of two cooperative agents. 
The \textit{Questioner} monitors recognition performance and generates context-aware prompts that highlight distinctions between easily confused actions. 
Additionally, the \textit{Selector} filters the multi-turn responses, extracting salient joints and discriminative cues that are injected into the model through local joint constraints and global semantic alignment. 
This agent-based interaction pipeline transforms targeted textual priors into training-aware guidance, enabling fine-grained action discrimination.

\subsection{Questioner: A Context-aware Prompter}
\label{sec:3_3}

Existing methods typically query the LLM with the single-turn static prompt, which often yields nearly identical descriptions for similar actions.
To tackle this issue, SkeletonAgent augments this paradigm with a \textit{Questioner} agent. 
Specifically, the \textit{Questioner} provides the LLM with a coarse-to-fine prompt chain to obtain an initial class-specific description for each action.
Then, the agent periodically profiles the class-wise recognition performance and identifies those hard-to-distinguish classes during training.
Accordingly, it constructs targeted prompts that guide the LLM toward context-aware discriminative outputs.

\textbf{Class-specific Description Preparation.}
As shown in \cref{fig:figure2} (a), SkeletonAgent initially asks the LLM to describe the intrinsic characteristics of each action class.
In practice, humans tend to build a global understanding of the action, then curiously zoom in to explore specific joints in greater detail. 
Inspired by this human perceptual routine, we design a coarse-to-fine prompting sequence that first requests a holistic overview of the motion, then successively focuses on the participating body parts, the most informative joints, and finally the key discriminative cues.
Compared to existing studies, our step-wise prompt chain can yield richer and more informative skeleton-based descriptions.
The complete prompts and sample responses construct the detailed class-specific descriptions for initial action perception, facilitating subsequent interactions.

\textbf{Performance Feedback Collection.}
Our target is to establish the interactive mechanism between recognition models and language models.
Therefore, we need to collect the real-time recognition feedback as guiding signals.
In practice, we maintain an online confusion matrix $\mathbf{H}(\cdot\,; t)$ to identify the current challenging actions for each class (\ie, similar-class set) at epoch $t$.

Specifically, the $(m,n)$-th entry denoted as $\mathbf{H}(m,n\,;t)$, counts the samples of class $m$ that are misclassified as belonging to class $n$.
Here, $m,n\in\{1,\dots, N_{c}\}$, and $N_{c}$ is the total number of action classes. 
Therefore, we define the similar-class set $\mathcal{S}_k(m)$ for class $m$ as the $k$ classes with the highest mis-classification counts in the current epoch
\begin{equation}
    \mathcal{S}_k(m)=\left\{\, n \;\middle|\; n \in \operatorname{Top}_k\!\bigl(\mathbf{H}(m,\cdot)\bigr) \right\}
\end{equation}
where $\operatorname{Top}_k(\mathbf{H}(m,\cdot))$ returns the indices of the $k$ largest entries in the $m$-th row of the mis-classification matrix $\mathbf{H}$.
We employ the periodic setting to enhance efficiency and avoid insufficient learning in the early stages.
Essentially, $\mathcal{S}_k(\cdot)$ captures the most confusable actions for each class.
The advantage of confusion matrices lies in their capacity to provide the most genuine recognition feedback, which may differ from the predictions of human visual mechanisms.
Consequently, the collected performance feedback can provide more valuable contextual information for training in subsequent epochs.

\begin{figure}[t]
\centering
\includegraphics[width=0.98\linewidth]{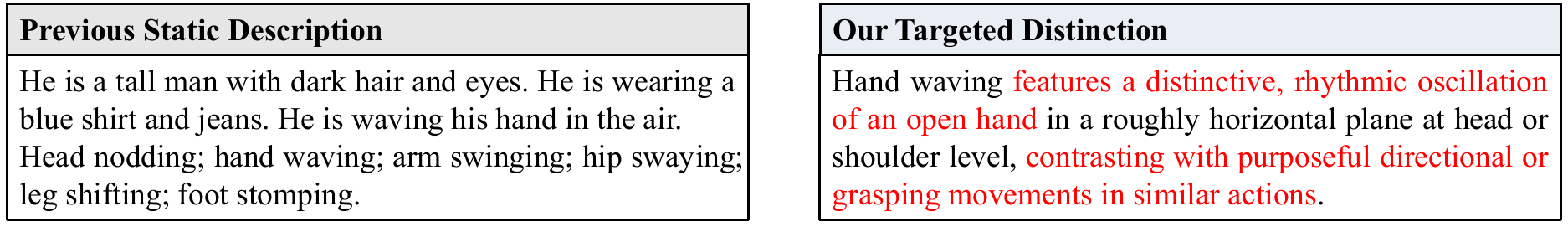}	
\caption{
Comparisons of static descriptions and our targeted distinctions for the \textit{hand waving} action.
The proposed approach captures distinctive cues and clearly emphasizes key characteristics that differ from other actions.
}
\label{fig:prompt}
\end{figure}

\textbf{Context-aware Prompt Generation.}
After constructing the similar list $\mathcal{S}$, we pass the current challenging actions of each class to the LLM via the targeted prompt shown in \cref{fig:figure2} (b).
This context-aware profile serves to encourage the LLM to focus on the subtle motion cues that distinguish the mentioned class from its most easily confused actions, thereby highlighting its targeted distinctive characteristics in the corresponding responses.

According to the results in \cref{fig:prompt}, the static description from GAP~\cite{xiang2023generative} actually lacks detailed information on key characteristics for the \textit{hand waving} action.
In contrast, our approach can capture more salient cues and clearly differentiate them from similar actions.
After prolonged training, the framework can leverage interaction history to maintain consistent constraints and mitigate the effects of overfitting.
In this way, the \textit{Questioner} supplements the targeted descriptions with the class-wise confusion profile, thereby providing the LLM with explicit feedback on fine-grained discrimination.

\subsection{Selector: An Adaptive Knowledge Highlighter}
\label{sec:3_4}

The previous holistic aligning paradigm lacks the fine-grained control over the correspondence between textual fragments and motion patterns.
Meanwhile, the interactive mechanisms introduce more descriptions, where the truly informative knowledge needs to be refined to enable recognition models to comprehend these guidance cues.
Therefore, we propose the \textit{Selector} agent that summarizes critical joints and distinguishing characteristics in the conversation history.
The module converts the key information into local joint constraints and global semantic alignment that can be leveraged by recognition models.

\textbf{Local Joint Constraints.}
By parsing the LLM’s answer, the joints and body parts most salient for action recognition can be readily identified. 
This information is injected into the recognition model through an auxiliary topology matrix $\mathbf{K} \in \mathbb{R}^{V\times V}$ for each class, which serves as an explicit structural constraint.
Specifically, $\mathbf{K}$ is initialized to zeros, and for each joint $v$ identified as salient, the entire row $\mathbf{K}(v, \cdot)$ is set to ones, effectively broadcasting that joint’s influence to all others.
As shown in Fig.~\ref{fig:figure2} (b), the matrix reflects a set of the most noteworthy significant joints, serving to remind the recognition model during training to consistently focus on these key relationships.

To incorporate $\mathbf{K}$ into the recognition pipeline, we first pool the final output of the network, \ie, $\mathbf{X}^{(L)}\in \mathbb{R}^{C \times T \times V}$, along the temporal dimension $T$, yielding a preliminary joint–level representation $\mathbf{x}_{\text{pre}}\!\in\!\mathbb{R}^{C\times V}$. 
Afterward, the skeleton feature with critical joints emphasized, denoted as $\mathbf{x_{joint}}$, is obtained through
\begin{equation}
	\mathbf{x_{joint}} = {\rm Proj \,} (\mathbf{x_{pre}} \cdot \mathbf{K})
	\label{eq_explicit_1}
\end{equation}
where ${\rm Proj \,}(\cdot)$ is the projection operation consisting of a convolutional layer followed by reshaping.
To make the encoder attend more strongly to these highlighted joints, we add the explicit constraint loss
\begin{equation}
	\mathcal{L}_{con} = - \mathbf{y} \log \mathbf{p_{\theta}(x_{joint})}
	\label{eq_explicit_2}
\end{equation}
where $\mathbf{y}$ is the one-hot ground-truth label, and $\mathbf{p_{\theta}(x_{joint})}$ denotes the predicted class distribution.
This supervision directs the model to exploit the injected joint priors, thereby achieving more discriminative representations.

\textbf{Global Semantics Alignment.}
In addition to the explicit joint-level constraint, we employ the global semantic alignment. 
Unlike previous methods that pair the skeletal representation with a generic context-agnostic description, we align it with the embedding of the Q6 response in Fig.~\ref{fig:figure2} (b), \ie, text deliberately crafted to distinguish the target action from its most confusable counterparts.

Specifically, we adopt a bidirectional matching scheme to enhance the semantic alignment of the embeddings of skeleton and text.
We employ the pre-trained CLIP model~\cite{radford2021learning} as the text encoder.
The above action descriptions are input into the encoder, undergoing standard tokenization and transformation to yield the corresponding textual features.
Then, we define two probability distributions $\mathbf{p_{i}^{s2t}}$ and $\mathbf{p_{i}^{t2s}}$, which represent the transition probabilities in skeleton-to-text and text-to-skeleton directions:
\begin{align}
\mathbf{p_{i}^{s2t}(s_i)} = \frac{\exp(cos(\mathbf{s_i, t_i})/\tau)} {\sum_{j \in \mathcal{B}} \exp(cos(\mathbf{s_i, t_j})/\tau)} \label{eq_prob_1} \\
\mathbf{p_{i}^{t2s}(t_i)} = \frac{\exp(cos(\mathbf{t_i, s_i})/\tau)} {\sum_{j \in \mathcal{B}} \exp(cos(\mathbf{t_i, s_j})/\tau)} \label{eq_prob_2}
\end{align}
Here, $\mathcal{B}$ is the batch size, $i, j$ are the index of samples within the batch, $cos(\cdot)$ represents the cosine similarly, $\tau$ is temperature parameter, and $\mathbf{s, t}$ represent the skeletal and textual representations obtained from the skeleton encoder and pre-trained text encoder, respectively.

Considering that multiple skeleton samples correspond to the same class, there could be more than one positive pair within the same batch.
Therefore, we use KL divergence to measure the probability distributions between skeletal features and textual features, and the alignment loss $\mathcal{L}_{align}$ can be written as
\begin{equation}
  \mathcal{L}_{align} = \frac{1}{2} (KL(\mathbf{p^{s2t}(s), q^{s2t}}) + KL(\mathbf{p^{t2s}(t), q^{t2s}}))
\label{eq_KL}
\end{equation}
where $\mathbf{q^{s2t}}$ and $\mathbf{q^{t2s}}$ represent the ground-truth probability distributions, with the positive pair probability set to 1 and the negative pair probability set to 0.
Through this targeted semantic guidance, the recognition model can focus more on the key discriminative basis, thereby enhancing the synergistic skeleton-semantic information.

Finally, the overall training objective function can be written as
\begin{equation}
  \mathcal{L} = \mathcal{L}_{cls} + \alpha \mathcal{L}_{con} + \beta \mathcal{L}_{align}
  \label{eq_overall}
\end{equation}
where $\mathcal{L}_{cls}$ is the cross-entropy loss used to supervise the classification, and $\alpha, \beta$ are the weighting parameters to balance the importance of different loss terms.

\section{Experiments}

\subsection{Datasets}

\textbf{NTU RGB+D (NTU-60)}~\cite{shahroudy2016ntu} is a large-scale action recognition dataset that contains 56,880 indoor captured samples, performed by 40 subjects and classified into 60 classes. 
Each action sample is annotated with 25 skeleton joints and contains up to two subjects.
This dataset recommends two evaluation protocols: 
(1) cross-subject (X-Sub): train data are performed by 20 subjects, and test data are performed by the remaining 20 subjects. 
(2) cross-view (X-View): train data from camera views 2 and 3, and test data from camera view 1.

\noindent
\textbf{NTU RGB+D 120 (NTU-120)}~\cite{liu2019ntu} is an extended version of NTU RGB+D, and contains 114,480 samples over 120 classes.
The sequences are captured with 106 subjects and 32 distinct camera setups.
There are also two settings: 
(1) cross-subject (X-sub): samples from 53 subjects are used for training, while the remaining 53 subjects are used for testing. 
(2) cross-setup (X-set): train data comes from 16 even setup IDs, and test data comes from 16 odd setup IDs.

\noindent
\textbf{Kinetics-Skeleton}~\cite{kay2017kinetics} is derived from the Kinetics 400 video dataset, utilizing the pose estimation toolbox to extract 240,436 training samples and 19,796 evaluation skeleton samples across 400 classes. 
For each sequence, two people are selected for multi-person clips based on the average joint confidence.
We adopt the available data provided by PYSKL~\cite{duan2022pyskl}.
Following the standard evaluation protocol, the Top-1 and Top-5 accuracies are reported.

\noindent
\textbf{FineGYM}~\cite{shao2020finegym} is a fine-grained action recognition dataset with 29,000 videos of 99 gymnastic action classes.
Since the dataset contains diverse fine-grained actions, it requires recognition models to distinguish different sub-actions within the same video.
Therefore, this dataset poses greater challenges to current recognition methods.
We use the skeleton data provided by PYSKL~\cite{duan2022pyskl}.
The Top-1 accuracy is reported in the evaluation protocol.

\noindent
\textbf{UAV-Human}~\cite{li2021uav} is a large-scale human behavior understanding dataset with unmanned aerial vehicles, containing 155 classes from 119 subjects. 
The data are captured by drones in diverse environments.
It defines two evaluation protocols: cross-subject-v1 (CSv1) and cross-subject-v2 (CSv2), with 89 subjects for training and 30 for testing.
The subject IDs are different across protocols.
The classification accuracy is used for performance evaluation.

\subsection{Implementation Details}

Our implementation is conducted on a single RTX 3090 GPU with PyTorch.
We employ PYSKL~\cite{duan2022pyskl} as the skeleton extractor baseline. 
For the text encoder, we employ the pre-trained ViT-B/32 model of CLIP~\cite{radford2021learning} to extract semantic embeddings.
Concurrently, GPT-4o~\cite{hurst2024gpt} is utilized to generate interactive dialogues.
The model is trained for 150 epochs with the batch size set to 64.
The initial learning rate is set to 0.1 with a cosine learning rate scheduler.
The optimizer is SGD with a Nesterov momentum of 0.9 and a weight decay of $5\times10^{-4}$. 
The temperature $\tau$ is set to 0.1, and the upper threshold $k$ for class statistics is set to 10.
The query period of the framework is set to 5 epochs.
The weights $\alpha$ and $\beta$ are set to $0.2$ and $0.5$.
We adopt the widely adopted six-stream ensemble strategy proposed in~\cite{chi2022infogcn}.
In addition, we sample the skeleton sequences to 100 frames and follow the strategy used in~\cite{chen2021channel,duan2022pyskl,zhang2024shap} for data preprocessing.

\begin{table*}[t]
  \caption{
  Comparisons against state-of-the-art methods on the NTU RGB+D, NTU RGB+D 120, and Kinetics-Skeleton datasets in terms of classification accuracy (\%).
  }
  \label{tab:main}
  \centering
  \scriptsize
  \renewcommand\arraystretch{1.1}
  \setlength\tabcolsep{3.6pt}
  \begin{tabular}{l|c|c c|c c|c c}
    \toprule
    \multirow{2}{*}{Methods} & \multirow{2}{*}{Publication} 
    & \multicolumn{2}{c|}{NTU RGB+D} 
    & \multicolumn{2}{c|}{NTU RGB+D 120}
    & \multicolumn{2}{c}{Kinetics-Skeleton}
    \cr
    & & X-Sub & X-View & \, X-Sub & X-Set & \, Top-1 & Top-5 \cr
    \midrule
    ST-GCN \cite{yan2018spatial} & AAAI 2018 & 81.5 & 88.3 & 70.7 & 73.2 & 30.7 & 52.8 \\
	  2s-AGCN \cite{shi2019two} & CVPR 2019 & 88.5 & 95.1 & 82.5 & 84.2 & 36.1 & 58.7 \\
    MS-G3D \cite{liu2020disentangling} & CVPR 2020 & 91.5 & 96.2 & 86.9 & 88.4 & 38.0 & 60.9 \\
    MST-GCN \cite{chen2021multi} & AAAI 2021 & 91.5 & 96.6 & 87.5 & 88.8 & 38.1 & 60.8 \\
    CTR-GCN \cite{chen2021channel} & ICCV 2021 & 92.4 & 96.8 & 88.9 & 90.6 & - & - \\
    STF \cite{ke2022towards} & AAAI 2022 & 92.5 & 96.9 & 88.9 & 89.9 & 39.9 & - \\
    InfoGCN \cite{chi2022infogcn} & CVPR 2022 & 93.0 & 97.1 & 89.8 & 91.2 & - & - \\
    PYSKL \cite{duan2022pyskl} & ACM MM 2022 & 92.6 & 97.4 & 88.6 & 90.8 & 49.1 & - \\
    FR-Head \cite{zhou2023learning} & CVPR 2023 & 92.8 & 96.8 & 89.5 & 90.9 & - & - \\
    UPS \cite{foo2023unified} & CVPR 2023 & 92.6 & 97.0 & 89.3 & 91.1 & 40.5 & 63.3 \\
    GAP \cite{xiang2023generative} & ICCV 2023 & 92.9 & 97.0 & 89.9 & 91.1 & - & - \\
	  HD-GCN \cite{lee2023hierarchically} & ICCV 2023 & 93.4 & 97.2 & 90.1 & 91.6 & 40.9 & 63.5 \\
    DS-GCN \cite{xie2024dynamic} & AAAI 2024 & 93.1 & 97.5 & 89.2 & 91.1 & 50.6 & - \\
	  BlockGCN \cite{zhou2024blockgcn} & CVPR 2024 & 93.1 & 97.0 & 90.3 & 91.5 & - & - \\
    SkateFormer \cite{do2024skateformer} & ECCV 2024 & 93.5 & 97.8 & 89.8 & 91.4 & - & - \\
    ProtoGCN \cite{liu2025revealing} & CVPR 2025 & 93.8 & 97.8 & 90.9 & 92.2 & 51.9 & 75.6 \\
    Hyper-GCN \cite{zhou2025adaptive} & ICCV 2025 & 93.7 & 97.8 & 90.9 & 92.0 & - & - \\
    \midrule
    \cellcolor{gray!20}Ours (2-ensemble) & \cellcolor{gray!20} & \cellcolor{gray!20}{94.0} & \cellcolor{gray!20}{97.7} & \cellcolor{gray!20}{90.5} & \cellcolor{gray!20}{92.4} & \cellcolor{gray!20}{52.2} & \cellcolor{gray!20}{75.9} \\
    \cellcolor{gray!20}Ours (4-ensemble) & \cellcolor{gray!20} & \cellcolor{gray!20}{94.3} & \cellcolor{gray!20}{98.0} & \cellcolor{gray!20}{91.4} & \cellcolor{gray!20}{93.0} & \cellcolor{gray!20}{53.0} & \cellcolor{gray!20}{76.8} \\
    \cellcolor{gray!20}\textbf{Ours (6-ensemble)} & \cellcolor{gray!20} & \cellcolor{gray!20}\textbf{94.5} & \cellcolor{gray!20}\textbf{98.2} & \cellcolor{gray!20}\textbf{91.7} & \cellcolor{gray!20}\textbf{93.1} & \cellcolor{gray!20}\textbf{53.6} & \cellcolor{gray!20}\textbf{77.0} \\
    \bottomrule
  \end{tabular}
\end{table*}

\begin{table}[t]
\scriptsize
\noindent
\begin{minipage}[t]{0.48\textwidth}
\centering
\captionof{table}{
Comparisons against state-of-the-art methods on FineGYM in terms of classification accuracy (\%).
}
\label{tab:FineGYM}
\setlength{\tabcolsep}{4pt}{
\renewcommand\arraystretch{1.1}
\begin{tabular}{l c c}
\toprule
\multirow{2}{*}{Methods}
& \multirow{2}{*}{Year}
& \multicolumn{1}{c}{FineGYM} \cr
& & Top-1 \cr
\midrule
MS-G3D \cite{liu2020disentangling} & 2020 & 92.0 \\
PYSKL \cite{duan2022pyskl} & 2022 & 94.1 \\
SkeletonMAE \cite{yan2023skeletonmae} & 2023 & 91.8 \\
DS-GCN \cite{xie2024dynamic} & 2024 & 93.4 \\
VA-AR \cite{wei2025va} & 2025 & 92.8 \\
ProtoGCN \cite{liu2025revealing} & 2025 & 95.9 \\
\midrule
\cellcolor{gray!20}\textbf{Ours} & \cellcolor{gray!20} & \cellcolor{gray!20}\textbf{96.5}  \\
\bottomrule
\end{tabular}
}
\end{minipage}
\hfill
\begin{minipage}[t]{0.48\textwidth}
\centering
\captionof{table}{
Comparisons against state-of-the-art methods on UAV-Human in terms of classification accuracy (\%).
}
\label{tab:UAV-Human}
\setlength{\tabcolsep}{4pt}{
\renewcommand\arraystretch{1.1}
\begin{tabular}{l c c c}
\toprule
\multirow{2}{*}{Methods}
& \multirow{2}{*}{Year}
& \multicolumn{2}{c}{UAV-Human} 
\cr
& & CSv1 & CSv2 \cr
\midrule
Shift-GCN \cite{cheng2020skeleton} & 2020 & 38.0 & 67.0 \\
FR-AGCN \cite{hu2022forward} & 2022 & 44.0 & 69.5 \\
TD-GCN \cite{liu2023temporal} & 2023 & 45.4 & 72.9 \\
HDBN \cite{liu2024hdbn} & 2024 & 48.0 & 75.4 \\
AL-GCN \cite{miao2024adaptive} & 2024 & 48.8 & 74.0 \\
TDSN-GCN \cite{liu2025tdsn} & 2025 & 47.8 & 74.2 \\
\midrule
\cellcolor{gray!20}\textbf{Ours} & \cellcolor{gray!20} & \cellcolor{gray!20}\textbf{52.1} & \cellcolor{gray!20}\textbf{78.1}  \\
\bottomrule
\end{tabular}
}
\end{minipage}
\end{table}

\subsection{Comparison with State-of-the-Art Methods}

We compare ours with the state-of-the-art methods on five datasets, including NTU RGB+D, NTU RGB+D 120, Kinetics-Skeleton, FineGYM, and UAV-Human.
According to the results, our model consistently establishes state-of-the-art performance with a significant margin in all scenarios. 
As shown in~\cref{tab:main}, for the NTU-60 dataset, the model achieves the best accuracy of 94.5\% on X-Sub and 98.2\% on X-View.
Remarkably, our model excels in performance on the challenging NTU-120 dataset, surpassing Hyper-GCN~\cite{zhou2025adaptive} by 0.8\% on X-Sub and 1.1\% on X-Set.
As for Kinetics, SkeletonAgent outperforms the state-of-the-art method ProtoGCN~\cite{liu2025revealing} by 1.7\%. 
Additionally, our model achieves state-of-the-art performance with a large margin on the FineGYM dataset in~\cref{tab:FineGYM}.
This confirms the efficacy of our approach for fine-grained action discrimination.
As shown in~\cref{tab:UAV-Human}, the significant improvement of SkeletonAgent on the UAV-Human dataset demonstrates its powerful ability to model diverse and complex actions.
Experimental results demonstrate the superiority of our method.

\begin{table}[t]
\caption{
Ablation study on the contribution of each component in SkeletonAgent under the NTU-60 X-Sub setting with the joint modality.
}
\label{tab:each}
\centering
\scriptsize
\renewcommand\arraystretch{1.1}
\setlength\tabcolsep{5.5pt}
\begin{tabular}{c c c c c c c}
\toprule
\multirow{2}{*}{Baseline} 
& \multicolumn{2}{c}{Questioner} 
& \multicolumn{2}{c}{Selector} 
& \multirow{2}{*}{Params (M)}
& \multirow{2}{*}{Acc (\%)}
\cr
& Stage 1 & Stage 2 & $\mathcal{L}_{con}$ & $\mathcal{L}_{align}$ & \cr
\midrule
\checkmark & -- & -- & -- & -- & 1.62 & 91.2 \\
\checkmark & \checkmark & -- & --& -- & 2.40 & 91.5 \\
\checkmark & \checkmark & -- & \checkmark & -- & 2.42 & 91.9 \\
\checkmark & \checkmark & \checkmark & \checkmark & -- & 3.56 & 92.3 \\
\checkmark & \checkmark & \checkmark & -- & \checkmark & 3.72 & 92.6 \\
\cellcolor{gray!20}\checkmark & \cellcolor{gray!20}\checkmark & \cellcolor{gray!20}\checkmark & \cellcolor{gray!20}\checkmark & \cellcolor{gray!20}\checkmark & \cellcolor{gray!20}\textbf{3.75} & \cellcolor{gray!20}\textbf{92.8} \\
\bottomrule
\end{tabular}
\end{table}

\begin{table}[t]
\caption{
Comparison of classification accuracies (\%) based on (\textit{left}) existing multi-modal methods, (\textit{middle}) various LLMs, and (\textit{right}) different text encoders.
}
\label{tab:comparisons}
\centering
\scriptsize
\renewcommand\arraystretch{1.1}
\begin{minipage}{.32\linewidth}
  \centering
  \setlength\tabcolsep{4pt}
  \begin{tabular}{l c}
    \toprule
    Methods & Acc (\%) \cr
    \midrule
    Baseline & 91.2 \\
    w/ LA-GCN \cite{xu2025language} & 91.4 \\
    w/ GAP \cite{xiang2023generative} & 91.3 \\
    w/ PURLS \cite{zhu2024part} & 91.6 \\
    w/ Neuron \cite{chen2025neuron} & 91.4 \\
    \cellcolor{gray!20}Ours & \cellcolor{gray!20}\textbf{92.8} \\
    \bottomrule
  \end{tabular}
\end{minipage}
\hfill
\begin{minipage}{.32\linewidth}
  \centering
  \setlength\tabcolsep{4pt}
  \begin{tabular}{l c}
    \toprule
    LLMs & Acc (\%) \cr
    \midrule
    Gemini 2.5 & 92.2 \\
    Qwen 3 & 92.4 \\
    Claude Sonnet 4 & 92.5 \\
    GPT-5 & 92.6 \\
    DeepSeek-V3.1 & 92.6 \\
    \cellcolor{gray!20}GPT-4o & \cellcolor{gray!20}\textbf{92.8} \\
    \bottomrule
  \end{tabular}
\end{minipage}
\hfill
\begin{minipage}{.32\linewidth}
  \centering
  \setlength\tabcolsep{4pt}
  \begin{tabular}{l c}
    \toprule
    Text encoders & Acc (\%) \cr
    \midrule
    XFMR-16 & 92.0 \\
    XFMR-32 & 92.2 \\
    BERT & 92.4 \\
    CLIP-RN101 & 92.3 \\
    CLIP-ViT-B/16 & 92.6 \\
    \cellcolor{gray!20}CLIP-ViT-B/32 & \cellcolor{gray!20}\textbf{92.8} \\
    \bottomrule
  \end{tabular}
\end{minipage}
\end{table}

\begin{table}[t]
\caption{
Comparison of classification accuracies (\%) on three subsets for NTU-60 dataset and four challenging action classes.
}
\label{tab:similar_classes}
\centering
\scriptsize
\renewcommand\arraystretch{1.1}
\setlength\tabcolsep{4pt}
\begin{tabular}{l c c c c}
  \toprule
  Methods & CTR-GCN \cite{chen2021channel} & FR-Head \cite{zhou2023learning} & ProtoGCN \cite{liu2025revealing} & Ours \\
  \midrule
  NTU-60 Hard   & 65.1 & 66.8 & 69.6 & \textbf{73.9\textsuperscript{\textcolor{darkred}{$\uparrow$4.3}}} \\
  NTU-60 Medium & 83.0 & 83.4 & 85.1 & \textbf{87.6\textsuperscript{\textcolor{darkred}{$\uparrow$2.5}}} \\
  NTU-60 Easy   & 95.1 & 95.6 & 95.9 & \textbf{96.6\textsuperscript{\textcolor{darkred}{$\uparrow$0.7}}} \\
  \midrule
  Play with phone & 69.1 & 73.8 & 75.3 & \textbf{78.2\textsuperscript{\textcolor{darkred}{$\uparrow$2.9}}} \\
  Writing         & 56.6 & 59.2 & 63.2 & \textbf{68.0\textsuperscript{\textcolor{darkred}{$\uparrow$4.8}}} \\
  Wear a shoe     & 79.5 & 78.8 & 85.7 & \textbf{93.4\textsuperscript{\textcolor{darkred}{$\uparrow$7.7}}} \\
  Take off a shoe & 75.6 & 77.7 & 83.8 & \textbf{88.3\textsuperscript{\textcolor{darkred}{$\uparrow$4.5}}} \\
  \bottomrule
\end{tabular}
\end{table}

\subsection{Ablation Study}

\noindent
\textbf{Effectiveness of Each Module.}
To verify the effectiveness of SkeletonAgent, we gradually add each component to conduct comparative experiments in \cref{tab:each}.
Specifically, the \textit{Questioner} can be divided into two stages: Stage 1 (class-specific description preparation) and Stage 2 (context-aware generation).
According to the results, we observe a performance improvement of 0.3\% with the class-specific description, and the explicit constraint $\mathcal{L}_{con}$ contributed an additional 0.4\% improvement.
Subsequently, we examine the contextual generation and semantics alignment $\mathcal{L}_{align}$, achieving improvements by 0.4\% and 0.7\%, respectively.
Since model inference requires only the skeleton encoder, the computational overhead introduced by the agents remains acceptable with the significant improvements.
By integrating all components, we surpass baseline performance by 1.6\%, which demonstrates the effectiveness of the proposed framework.

\noindent
\textbf{Effectiveness of Contextual Interaction.}
To examine the efficacy of the dynamic interaction mechanism, we analyze the proposed SkeletonAgent and existing LLM-based approaches, including LA-GCN~\cite{xu2025language}, GAP~\cite{xiang2023generative}, PURLS~\cite{zhu2024part}, and Neuron \cite{chen2025neuron}.
In practice, we employ the static descriptions and human expert annotations provided by these methods.
As shown in \cref{tab:comparisons}, while they facilitate the differentiation, the improvement of our method is more significant.
The key to interaction lies in knowing what the recognition model truly requires.
Therefore, the proposed approach offers superior advantages over static descriptions and fixed expert annotations.
The results indicate that the agent-based dynamic interaction mechanism can effectively enhance performance.

\noindent
\textbf{Influences of LLMs and Text Encoders.}
We compare different LLMs for interactions in \cref{tab:comparisons}.
Among these models, GPT-4o could provide more discriminative guidance by using the proposed interactive agents, and the generated descriptions also achieve the best performance.
The results demonstrate the robustness of SkeletonAgent when using different LLMs, with performance differences less than 0.6\%.
We also employ additional constraints and verification designs in the interactive prompts to prevent potential hallucinations in LLM outputs.
Detailed instructions are provided in the supplementary materials.
The results show that the text encoder from CLIP achieves the best performance.
Therefore, we use CLIP-ViT-B/32 as the default text encoder.

\begin{figure}[t]
\centering
\begin{subfigure}[b]{0.24\linewidth}
\includegraphics[width=\linewidth]{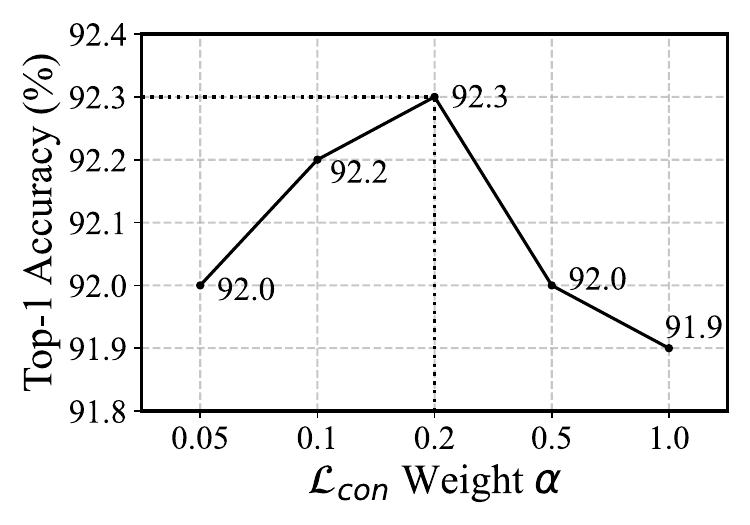}
\caption{Accuracy vs. $\alpha$}
\label{fig:short-a}
\end{subfigure}
\hfill
\begin{subfigure}[b]{0.24\linewidth}
\includegraphics[width=\linewidth]{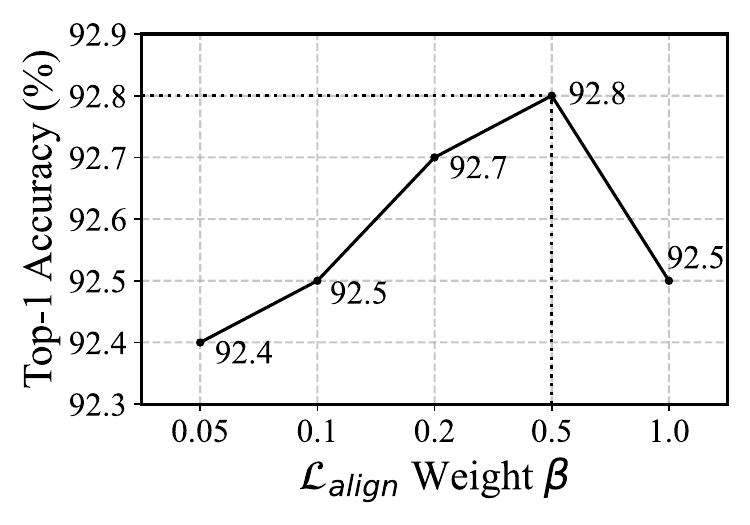}
\caption{Accuracy vs. $\beta$}
\label{fig:short-b}
\end{subfigure}
\hfill
\begin{subfigure}[b]{0.48\linewidth}
\includegraphics[width=\linewidth]{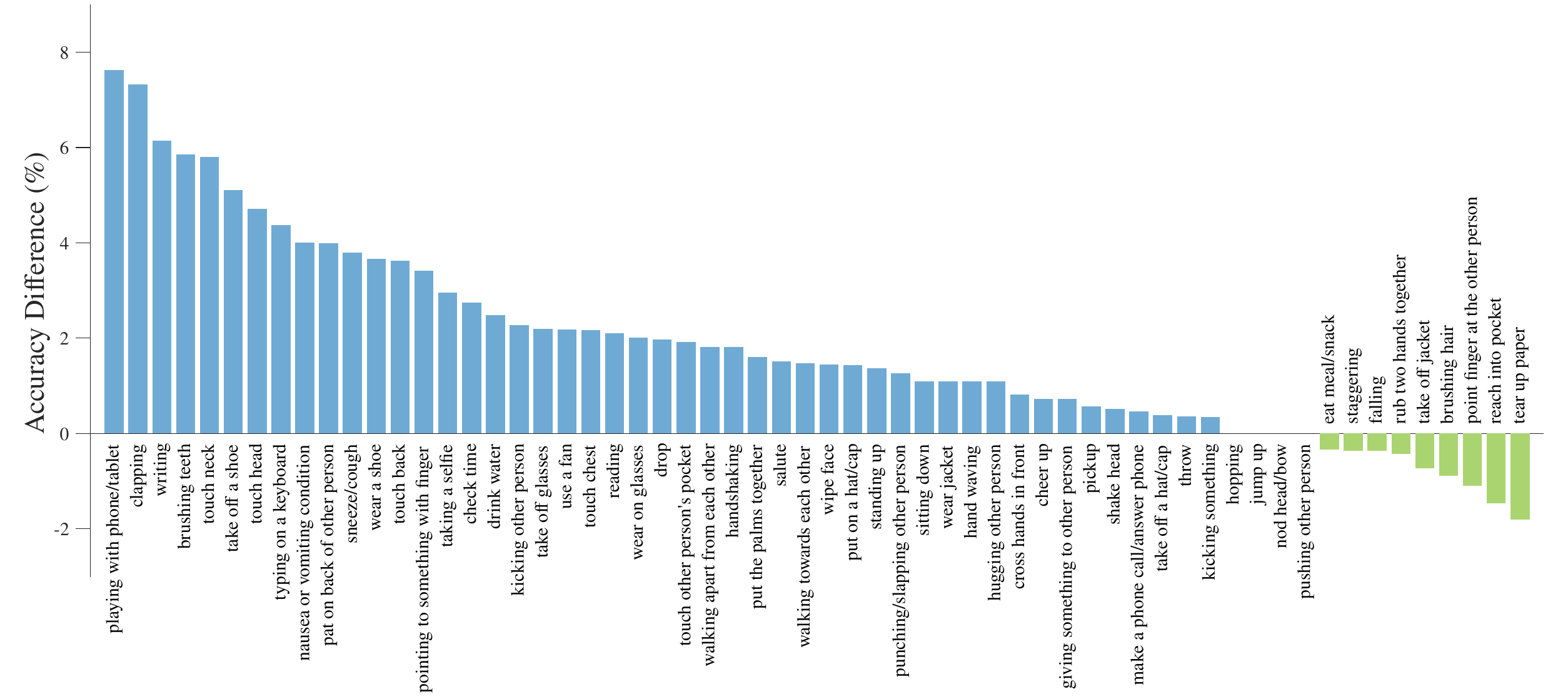}
\caption{Accuracy difference with the baseline}
\label{fig:short-c}
\end{subfigure}
\caption{
Ablation on the influences of the weights and the accuracy difference (\%) with the baseline PYSKL~\cite{duan2022pyskl} under the NTU-60 X-Sub setting with the joint modality.
}
\label{fig:short}
\end{figure}

\begin{figure}[t]
\centering
\subfloat[Epoch 10]
{\includegraphics[width=.24\linewidth]{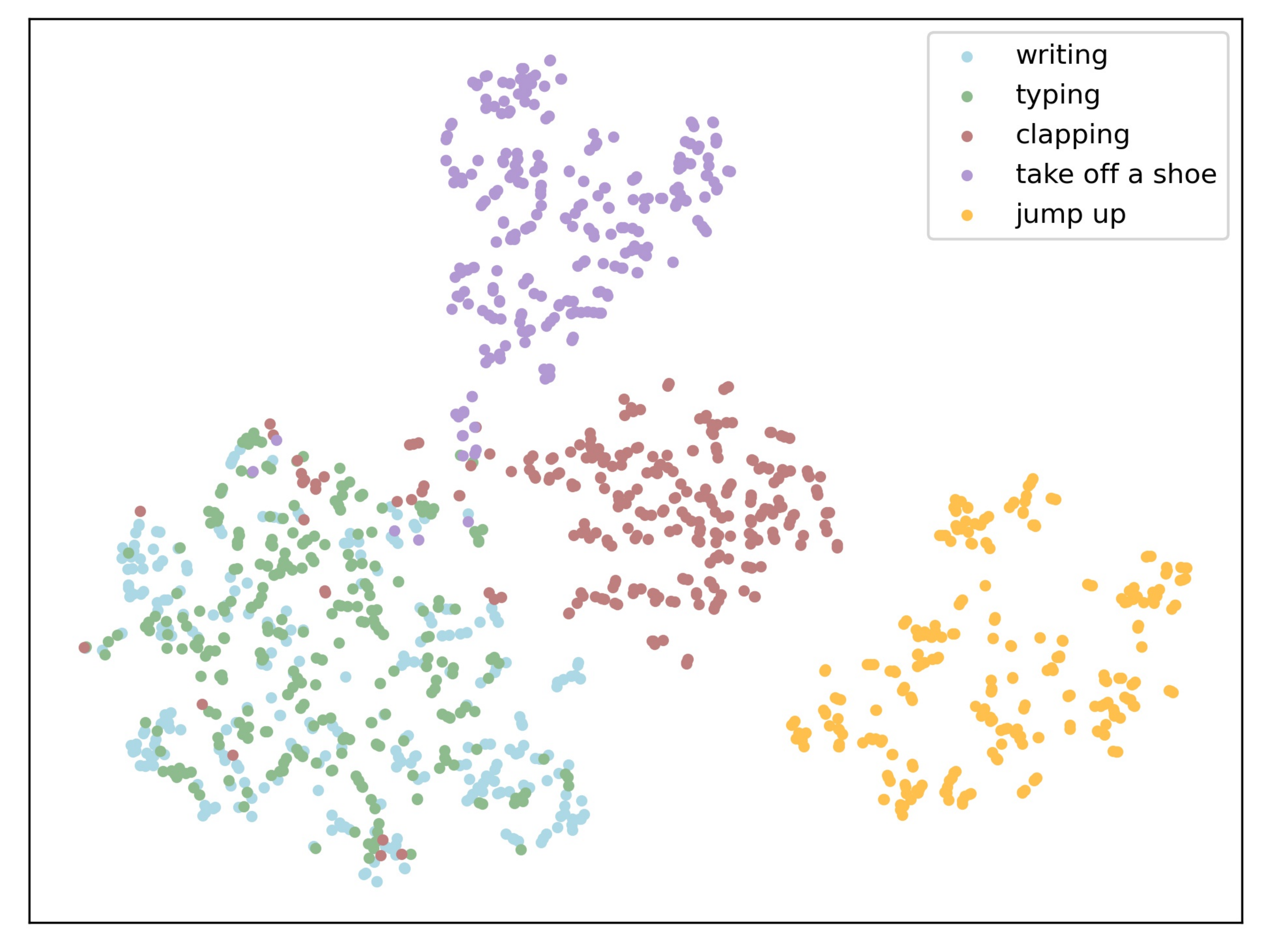}\label{fig_6_1}}
\hfill
\subfloat[Epoch 50]
{\includegraphics[width=.24\linewidth]{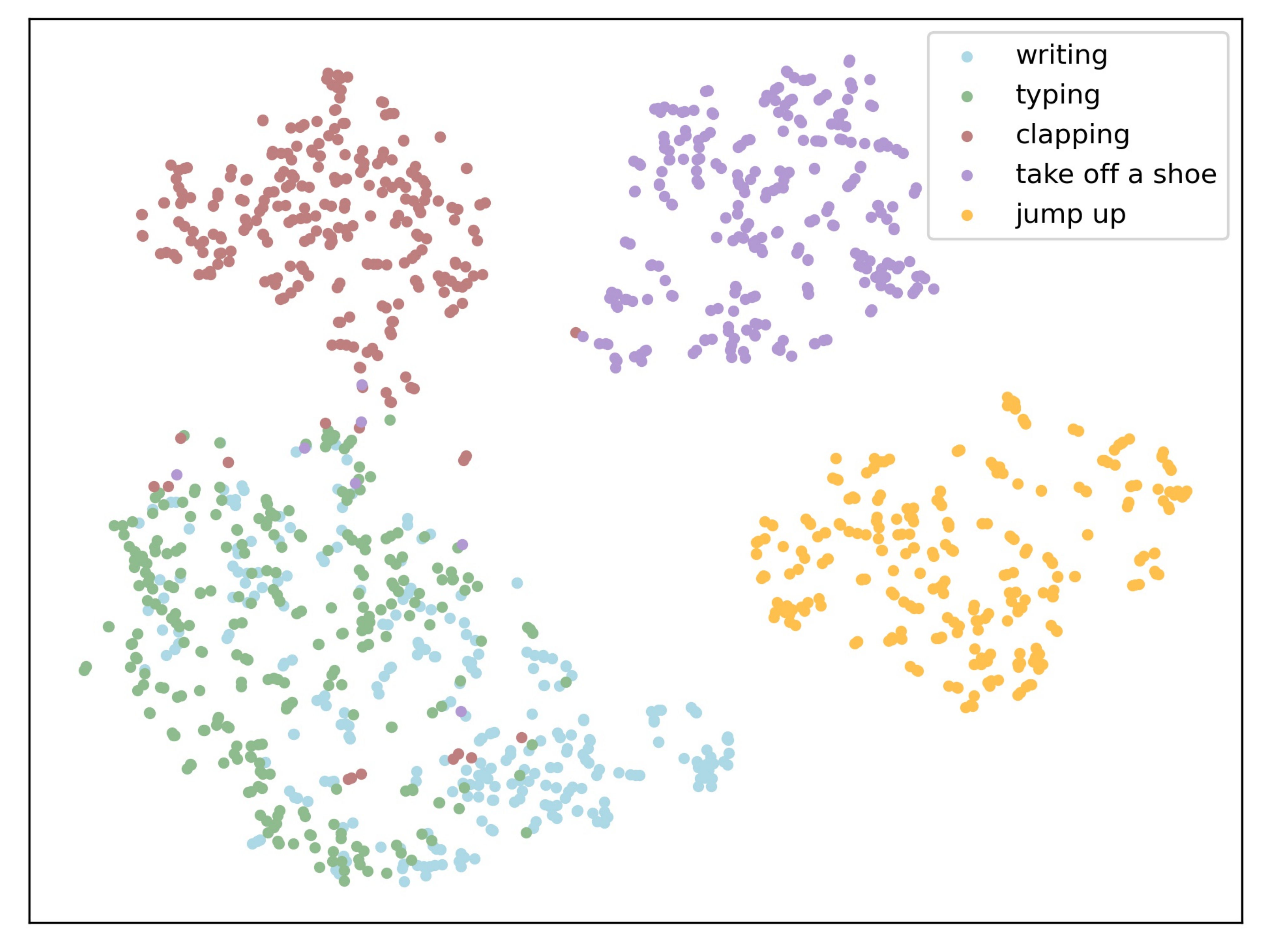}\label{fig_6_2}}
\hfill
\subfloat[Epoch 100]
{\includegraphics[width=.24\linewidth]{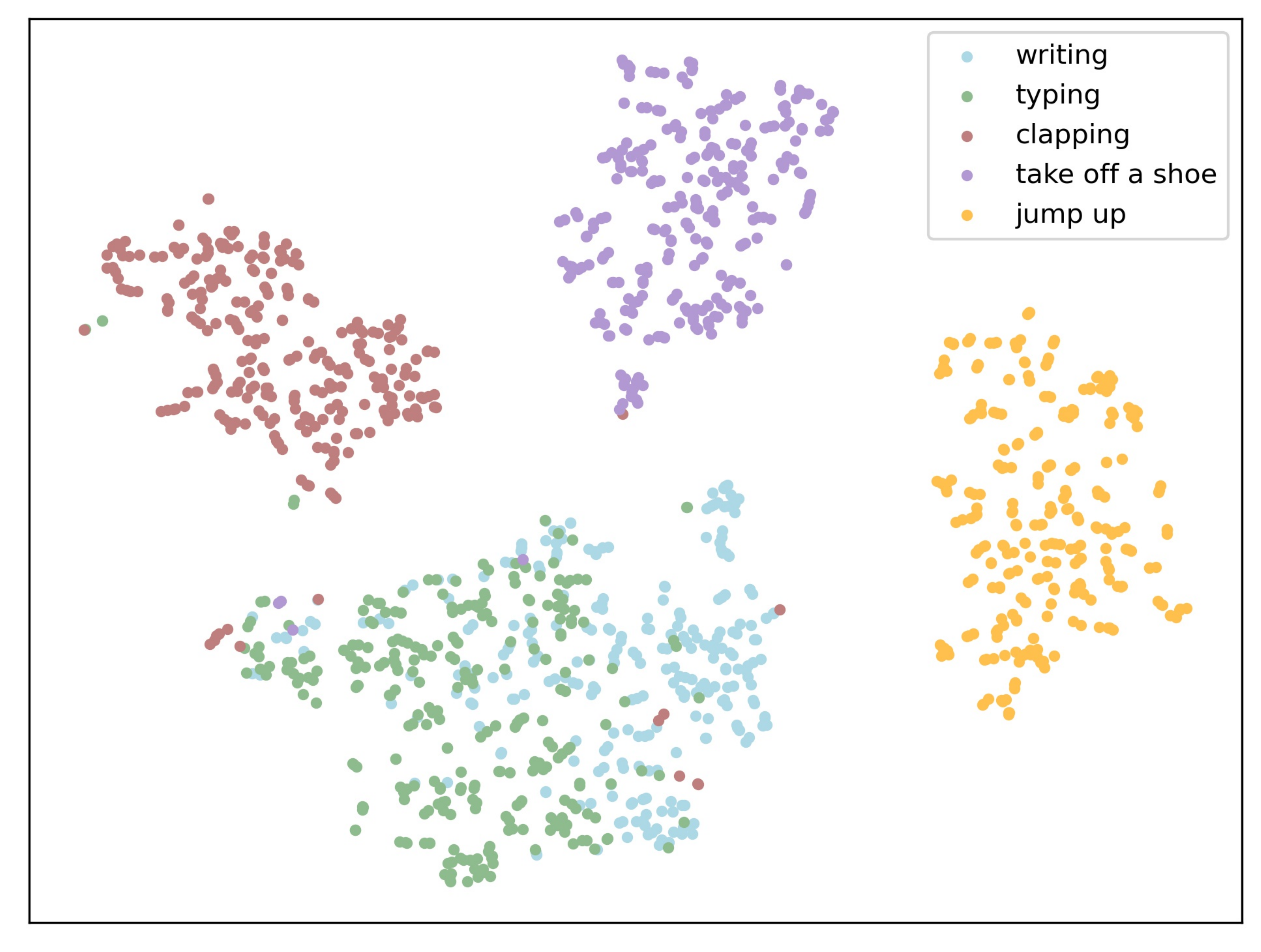}\label{fig_6_3}}
\hfill
\subfloat[Epoch 150]
{\includegraphics[width=.24\linewidth]{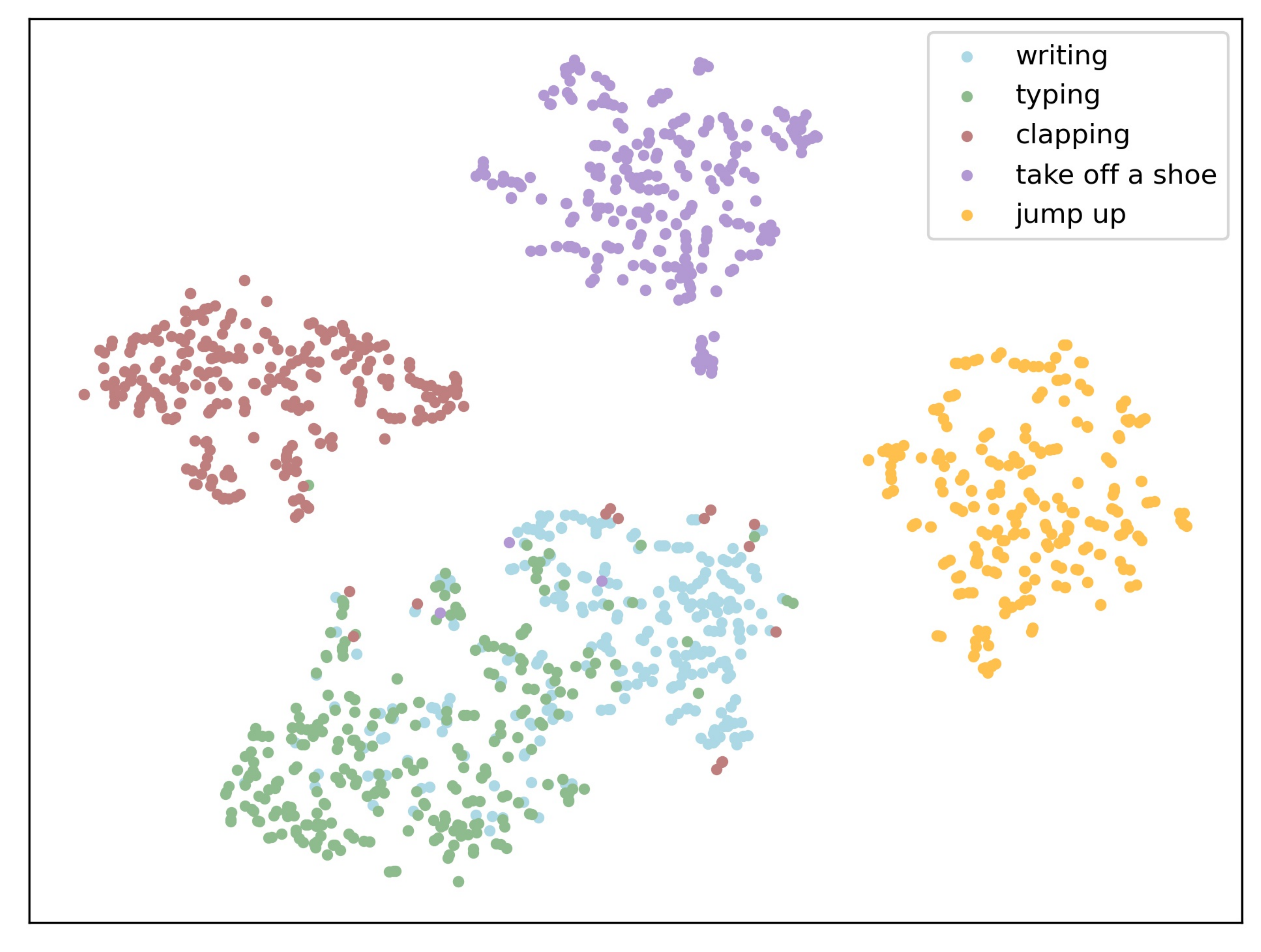}\label{fig_6_4}}
\caption{
The t-SNE visualization of the feature embeddings for five action classes from NTU-60 throughout the training process. 
Best viewed with zoom in.
}
\label{fig:t-SNE}
\end{figure}

\noindent
\textbf{Effect of Hyper-parameters.}
We analyze the effect of the weights $\alpha$ and $\beta$ in \cref{fig:short}.
The value for $\alpha$ is determined to be 0.2, corresponding to peak accuracy.
We also explore combinations of varying $\beta$ to balance explicit constraint and implicit alignment.
The best performance is achieved when $\beta$ is equal to 0.5.

\noindent
\textbf{Performance on Similar Classes.}
To demonstrate the superiority in distinguishing similar actions, we present the detailed comparisons in \cref{tab:similar_classes}.
According to the results of CTR-GCN~\cite{chen2021channel}, we split the NTU-60 dataset into three subsets based on difficulty and computed the average accuracy.
Specifically, we gather actions whose accuracy is lower than 70\% as Hard set, between 70\% and 90\% as Medium set, and over 90\% as Easy set. 
We compare our method with two other approaches designed to refine similar actions.
The experiment is also under the X-Sub setting with the joint modality. 
The results show that SkeletonAgent consistently improves performance, especially in classes with greater difficulties.
The following four rows demonstrate the significant gains achieved by the proposed method on four challenging actions.
In addition, \cref{fig:short-c} presents the accuracy difference between our method and PYSKL on the NTU-60 dataset.
From the comparisons, we can observe that SkeletonAgent offers more pronounced improvements in a significantly greater number of classes.
We also plot the t-SNE visualization of skeleton representation distribution over epochs in \cref{fig:t-SNE}.
The visualization reflects the guiding effect of refined textual semantics on skeletal representations.
The results can further validate the effectiveness of the proposed framework in fine-grained discrimination capabilities.

\begin{figure}[t]
\centering
\includegraphics[width=\linewidth]{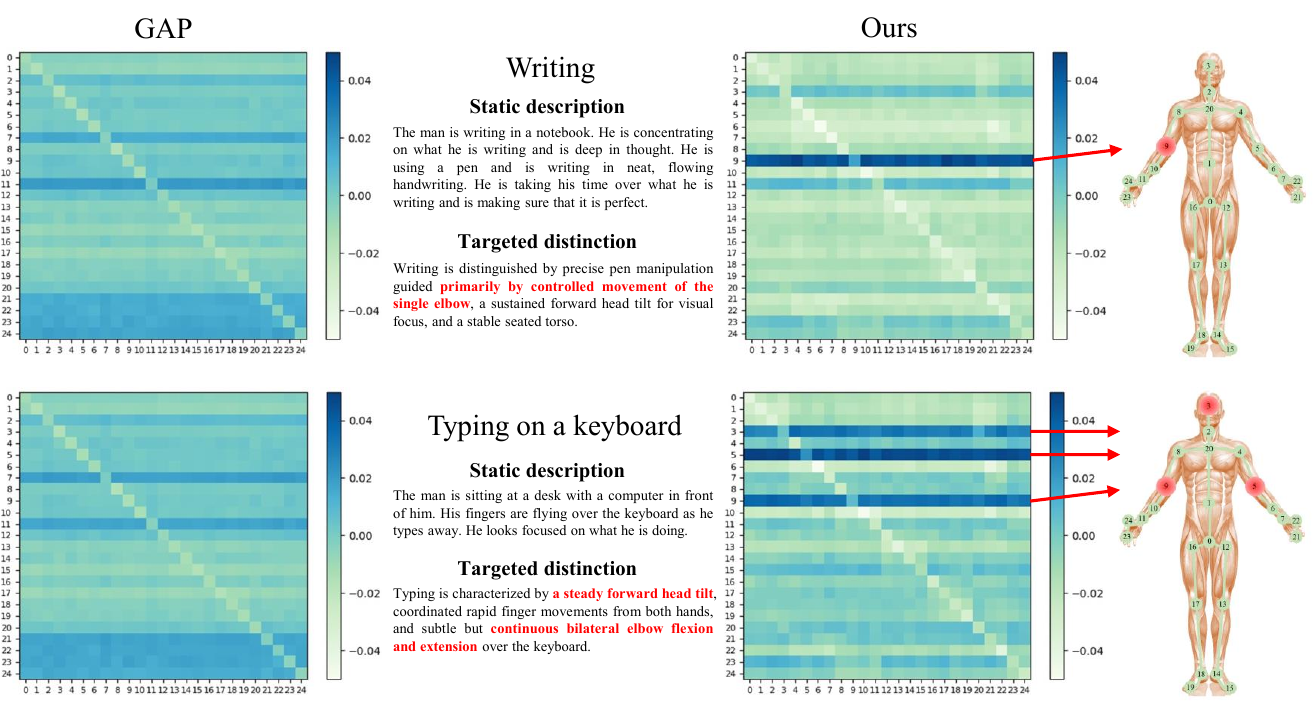}
\caption{ 
Visualization of action descriptions and the topologies learned by GAP~\cite{xiang2023generative} and our method for similar actions \textit{Writing} and \textit{Typing on a Keyboard}. 
Darker colors indicate the stronger correlation between corresponding joints.
}
\label{fig:figure5}
\end{figure}

\noindent
\textbf{Discussion.}
The key to our approach lies in the proposed interactive feedback mechanism, which reveals the information truly required by the recognition model and provides targeted guidance to focus on the most noteworthy distinctions.
Compared to previous static methods, our framework can achieve significant improvements with acceptable overhead.
The detailed analysis of computational costs can be found in the supplementary materials.

\noindent
\textbf{Visualization.}
To visually demonstrate the effectiveness of SkeletonAgent, we present the related descriptions and the topologies for two similar actions \textit{writing} and \textit{typing on a keyboard} in \cref{fig:figure5}.
The static descriptions are provided by GAP~\cite{xiang2023generative}, and the topologies are shown by averaging $\mathbb{R}^{N\times N\times C}$ along the $C$ dimension.
It is observed that previous single-round descriptions lack sufficient context, causing the model to remain constrained to similar hand motions.
In contrast, our targeted distinctions leverage dynamic interactions to explicitly highlight key features, guiding the model to focus on more discriminative single-arm and double-arm motions.
The color patterns within topologies reflect that the proposed interaction framework enables the recognition model to effectively focus on crucial differences associated with target motion.

\section{Conclusion}

In this paper, we introduced SkeletonAgent, a novel interactive framework for skeleton-based action recognition that leverages LLM agents to enable dynamic collaboration between recognition models and language models.
Through this multi-turn iterative mechanism, our approach generates targeted discriminative guidance for challenging actions based on continuous feedback from recognition models.
The interactive process allows the framework to progressively identify and emphasize crucial fine-grained information, thereby achieving more precise cross-modal alignment between skeletal and semantic representations.
Extensive experiments on five benchmark datasets demonstrate the effectiveness of SkeletonAgent, which establishes new state-of-the-art performance across the board.

%
%
\bibliographystyle{splncs04}
\bibliography{main}
\end{document}